\documentclass[11pt,a4paper]{article}
\usepackage[hyperref]{emnlp-ijcnlp-2019}
\usepackage{times}
\usepackage{latexsym}

\usepackage{url}

\usepackage{amsfonts}
\usepackage{amsmath}
\usepackage{amssymb}
\usepackage{graphicx}
\usepackage[shortlabels]{enumitem}
\usepackage{booktabs}
\usepackage{multirow}

\aclfinalcopy 

\newcommand\sS{\mathcal{S}}
\newcommand\sZ{\mathcal{Z}}

\newcommand\obqa{\textsc{OpenBookQA}}
\newcommand\csqa{\textsc{CommonsenseQA}}
\newcommand\mnli{\textsc{MNLI}}

\title{Are We Modeling the Task or the Annotator? An Investigation of Annotator Bias in Natural Language Understanding Datasets}

\author{Mor Geva \\
  Tel Aviv University, \\
  Allen Institute for AI \\
  {\tt morgeva@mail.tau.ac.il} \\\And
  Yoav Goldberg \\
  Bar-Ilan University, \\
  Allen Institute for AI \\
  {\tt yoav.goldberg@gmail.com} \\\AND
  Jonathan Berant \\
  Tel Aviv University, \\
  Allen Institute for AI \\
  {\tt joberant@cs.tau.ac.il} \\}
  
\date{}

\begin{document}
\maketitle

\begin{abstract}
Crowdsourcing has been the prevalent paradigm for creating natural language understanding datasets in recent years.
A common crowdsourcing practice is to recruit a small number of high-quality workers, and have them massively generate examples.
Having only a few workers generate the majority of examples raises concerns about data diversity, especially when workers freely generate sentences.
In this paper, we perform a series of experiments showing these concerns are evident in three recent NLP datasets.
We show that model performance improves when training with annotator identifiers as features, and that models are able to recognize the most productive annotators.
Moreover, we show that often models do not generalize well to examples from annotators that did not contribute to the training set. 
Our findings suggest that annotator bias should be monitored during dataset creation, and that test set annotators should be disjoint from training set annotators.
\end{abstract}
\section{Introduction}

Generating large datasets has become one of the main drivers of progress in natural language understanding (NLU). The prevalent method for creating new datasets is through crowdsourcing, where examples are generated by workers \cite{zaidan2011crowdsourcing,richardson2013mctest,bowman2015large,rajpurkar2016squad,trischler2017newsqa}.
A common recent practice is to choose a small group of workers who produce high-quality annotations, and massively generate examples using these workers.

Having only a few workers annotate the majority of dataset examples raises concerns about data diversity and the ability of models to generalize, especially when the crowdsourcing task is to generate free text.
If an annotator consistently uses language patterns that correlate with the labels, a neural model can pick up on those, which can lead to an over-estimation of model performance.



In this paper, we continue recent efforts to understand biases that are introduced during the process of data creation \cite{levy2015supervised,schwartz2017roc,gururangan2018annotation, glockner2018breaking, poliak2018hypothesis, tsuchiya2018performance,aharoni2018split, paun2018comparing}.
We investigate this form of bias, termed \emph{annotator bias}, and perform multiple experiments over three recent NLU datasets: \mnli{} \cite{N18-1101}, \obqa{}. \cite{mihaylovetal2018}, and \csqa{} \cite{talmor2019commonsenseqa}. 

First, we establish that annotator information improves model performance by supplying annotator IDs as part of the input features. Second, we show that models are able to recognize annotators that generated many examples, illustrating that annotator information is captured by the model. Last, we test whether models generalize to annotators that were not seen at training time. We observe that often generalization to new annotators fails, and that augmenting the training set with a small number of examples from these annotators substantially increases performance.




Taken together, our experiments show that annotator bias exists in current NLU datasets, which can lead to problems in model generalization to new users.
Hence, we propose that annotator bias should be monitored
at data collection time 
and to tackle it by having the test set include examples from a disjoint set of annotators.

\section{Crowdsourcing Practice}

Crowdsourcing has become the prominent paradigm for creating NLP datasets \cite{callison2015crowdsourcing, sabou2014corpus}. It has been used for various NLU tasks, including Question Answering \cite{rajpurkar2018squad, mihaylovetal2018, dua2019drop}, commonsense and visual reasoning \cite{talmor2019commonsenseqa, zellers19, suhr2018corpus}, and Natural Language Inference (NLI) \cite{N18-1101}.

In a typical process, annotators are recruited and screened \cite{sabou2014corpus}, often resulting in a small group that creates most of the dataset examples.
\newcite{mihaylovetal2018} recruited a few dozens of qualified workers that wrote 5,957 questions.
\newcite{suhr2018corpus} recruited 99 workers for a sentence writing task, who created more than 100,000 examples.  \newcite{N18-1101} recruited 387 workers for writing over 400,000 sentences, while \citet{krishna2017visual} had 33,000 workers contributing 1.7 million examples.

These examples demonstrate that datasets are often constructed using a small number of annotators, approximately 1 annotator per $10^2$--$10^3$ examples.
Furthermore, (see Section~\ref{sec:setup}), the annotator distribution is skewed with a few annotators creating the vast majority of the dataset.
In tasks that involve creative language writing, this may have implications on data diversity, and lead to an over-estimation of model performance.

\section{Experimental Setup}
\label{sec:setup}

We focus on crowdsourcing tasks where workers produce full-length sentences.
We first describe the datasets we test our hypothesis on, and then
provide details on the model and training.

\paragraph{Datasets}
\label{sec:datasets}

We consider recent NLU datasets, for which the annotator IDs are available.
\begin{itemize}[topsep=4pt, itemsep=1pt, leftmargin=.2in, parsep=2pt]
    \item \textbf{\mnli{}} (matched) \cite{N18-1101}: A NLI dataset. Each example was created by introducing an annotator with a premise sentence and asking her to write a hypothesis sentence that is either entailed, contradicted or is neutral with respect to the premise.
    \item \textbf{\obqa{}} \cite{mihaylovetal2018}: A multiple-choice question answering dataset, focusing on multi-hop reasoning. Each question and its answer distractors were written by a worker, based on a given scientific fact.
    \item \textbf{\csqa{}} \cite{talmor2019commonsenseqa}: A multiple-choice question answering dataset, focused on commonsense knowledge. Questions were written by crowdworkers, who try to bridge between two concepts extracted from \textsc{ConceptNet} \cite{speer2017conceptnet}
\end{itemize}

\begin{table}[h]
    \centering
    \footnotesize
    \begin{tabular}{l|c|c}
          & \# examples & \# annotators \\
         \mnli{} (matched) & 402,517 & 380 \\
         \obqa{} & 5,457 & 84 \\
         \csqa{} & 11,096 & 132
         
    \end{tabular}
    \caption{Statistics for datasets used in our experiments.}
    \label{tab:datasets}
\end{table}

\begin{figure}[h]
\centering
\includegraphics[scale=0.44]{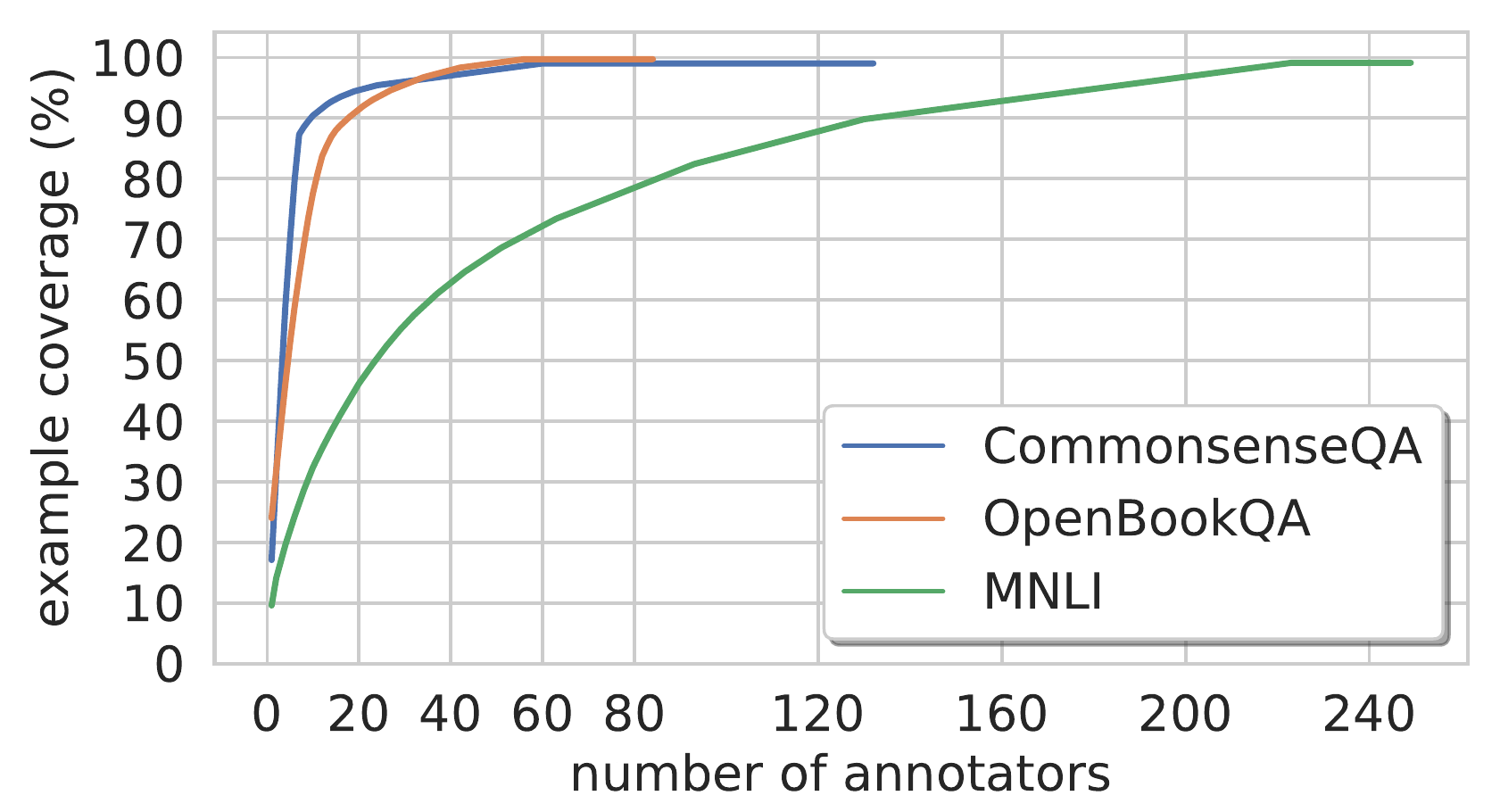}
\caption{Proportion of examples covered by number of annotators (sorted by number of annotations).}
\label{fig:example_coverage}
\end{figure}

Table~\ref{tab:datasets} summarizes the size and number of annotators who worked on each dataset. Figure~\ref{fig:example_coverage} shows the fraction of examples covered by the number of annotators, sorted by the number of examples they annotated. In all datasets, and specifically in \obqa{} and \csqa{}, most of the examples were generated by a small number of annotators.

\paragraph{Model}
\label{sec:model}
We use the pretrained BERT-base \cite{devlin2018bert}, a strong model obtaining close to state-of-the-art performance on all three datasets.  
We add a single linear layer over BERT outputs and apply the same fine-tuning procedure in all experiments: fine-tuning for 3 epochs, using batch size 10, learning rate $2\times 10^{-5}$, and maximum sequence length of 128 word pieces. 

\section{Experiments and Results}
We now describe a series of experiments for quantifying annotator bias, aiming to answer the following questions:
1) Do models perform better when exposed to the annotator ID?
2) Can models detect the annotators from their generated examples? 
3) Do models generalize across annotators?

\paragraph{The utility of annotator information}
Our first experiment aims to determine whether models perform better given \emph{perfect} information on the annotator ID.
To this end, we follow the standard way of feeding input to BERT \cite{devlin2018bert} and concatenate the annotator ID as an additional feature to every example in every dataset.
Formally, we replace every example $(x=(w_1, ..., w_{|x|}), y)$ created by annotator $z$, with the example $((z, w_1, ..., w_{|x|}), y)$, where $z$ is a textual  unique annotator identifier, $x$ is the input sequence and $y$ is the gold label.

We compare performance on the original datasets and their new version.
Adding the annotator ID improves model performance across all datasets (Table~\ref{tab:annotator_info}), 
showing that perfect annotator information is useful for prediction, and there is incentive for the model to capture this information.


\begin{table}[h]
    \centering
    \footnotesize
    \begin{tabular}{@{\hskip4pt}l@{\hskip4pt}|@{\hskip4pt}c@{\hskip4pt}|@{\hskip4pt}c@{\hskip4pt}||@{\hskip4pt}c@{\hskip4pt}}
          & Without ID & With ID & $p$-value\\
    \obqa{} & 52.2 & 56.4 & $1.83e^{-2}$ \\
    \csqa{} & 53.6 & 55.3 & $11.98e^{-2}$ \\
    \mnli{} & 82.9 & 84.5 & $5.13e^{-7}$ \\
    \end{tabular}
    \caption{Model development performance, after training with/without annotator IDs as additional inputs. Following \citet{dror2018hitchhiker}, $p$-values were calculated using the McNemar's test \cite{mcnemar1947note} for \mnli{} and the Bootstrap test \cite{berg2012empirical} for \obqa{} and \csqa{}.}
    \label{tab:annotator_info}
\end{table}

\paragraph{Annotator recognition}
Perfect annotator information improves performance, but it is possible that a model can recognize the annotators from the input sequence only, even without being exposed to the annotator ID explicitly. In the next experiment, we investigate the ability of models to recognize the annotators from the input.

To this end, we fine-tune BERT-base to predict annotator IDs from examples. We limit the task to 6 labels of the top-$5$ most productive annotators of each dataset and an \textsc{OTHER} label for all other annotators. Formally, we replace every example $(x, y)$ created by annotator $z$, with the example $(x, \bar{z})$, where $\bar{z} = z$ if $z$ is in the top-$5$ annotators and $\bar{z} = \text{OTHER}$ otherwise. 

\begin{figure}[ht]
\centering
\includegraphics[scale=0.45]{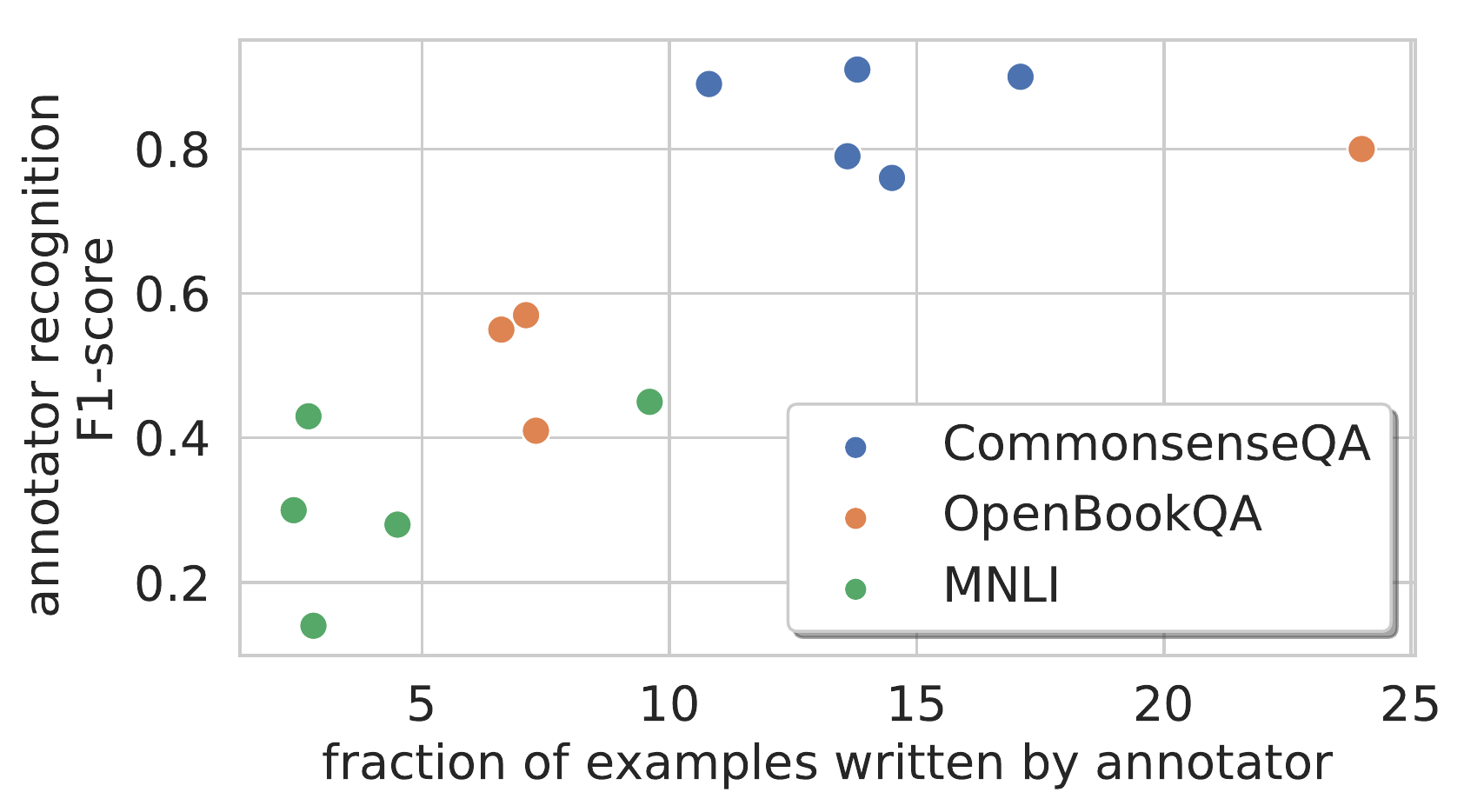}
\caption{Annotator recognition F1-scores for the top-$5$ annotators of each dataset. 
For \obqa{} only 4 data points are plotted, as the second annotator is not in the original development set.}
\label{fig:annotator_pred}
\end{figure}

Figure~\ref{fig:annotator_pred} shows the F1-score for the top-$5$ annotators of every dataset (y-axis), and the fraction of dataset examples created by each annotator (x-axis). Overall, annotators who write many examples are recognized better by the model: The model struggles to recognize \mnli{} annotators with F1 scores below $0.5$, and excels at recognizing annotators from \csqa{} with scores between $0.76$--$0.91$. For the top annotator of \obqa{}, who created 24\% of the dataset examples, the model obtains a high F1 score of $0.8$. To conclude the first two experiments, annotator ID information is useful for the downstream task, and also can be predicted with high probability from the input for a large fraction of the examples.

\paragraph{Generalization across annotators}
In our final experiment, we examine whether trained models generalize to new annotators. 
To address this question, we re-split each dataset, creating a training and development set with disjoint annotators:
Given a dataset with example set $\sS$, we denote by $\sS_z \subset \sS$ the subset of examples created by annotator $z$. Similarly, for a set of annotators $\sZ$, let $\sS_\sZ = \bigcup_{z \in \sZ}{\sS_{z}}$. 
We rank annotators by the number of examples they generated, and for each dataset $\mathcal{S}$, we create two types of data splits.  
For each annotator $z$ in the top-5 annotators, we create a \emph{single-annotator} data split with $\mathcal{S} \backslash \mathcal{S}_z$ and $\mathcal{S}_z$ as the train and development sets, respectively. Namely, we consider the examples created by annotator $z$ as the development set, while using all other examples for training.
In addition, we pick 5 sets of 5 annotators, who annotated a small number of examples, and for each such set $\sZ$, we create a 
\emph{multi-annotator} split with $\sS \backslash \sS_\sZ$ and $\sS_\sZ$ as the train and development sets, respectively.  Namely, we consider the examples created by the 5 annotators $\sZ$ as the development set, while using all other examples for training.
Overall, there are 5 single-annotator splits and 5 multi-annotator splits for each dataset. 


We fine-tune BERT-base and evaluate it on the development set, and compare the results to a random data split of identical size. We repeat every experiment 3 times, except for multi-annotator experiments on \obqa{} and \csqa{} which we repeat 9 times due to high variance.
Table~\ref{tab:annotator_versus_random_splits} shows the mean and standard deviation of performance difference (p.d.) between each annotator(s) split and its corresponding random split, where negative numbers indicate that performance on the annotator split was lower.

Our clearest finding is that in \obqa{} performance on the the multi-annotator split is dramatically lower than on a random split in all 5 annotator sets, where performance drops by up to 23 accuracy points. This shows that the model does not generalize to examples generated by unseen annotators. 
In the other datasets, results on the multi-annotator split are more varied, where performance drops in roughly half the cases, sometimes substantially -- up to 10 accuracy points in \csqa{} and 5 in \mnli{}.

In the single-annotator splits, in roughly half the cases performance on the annotator split was lower than the random split. However,
measuring p.d. only for single annotators might be misleading, because specific annotators vary in the difficulty of examples they produce. Thus, running a model on a new annotator that produces easy examples will not result in decreased performance. Next, we propose a more fine-grained experiment that controls for these two confounding factors.

\begin{table}[h]
    \setlength\tabcolsep{4pt}
    \centering
    \footnotesize
    \begin{tabular}{cc|cc}
          \multicolumn{2}{c}{\csqa{}-single} & \multicolumn{2}{c}{\csqa{}-multi} \\ 
          \toprule 
          $4.2 \pm 0.7$ & 17.1\% & $-9.5 \pm 8.3$ & 0.9\% \\ 
          $7.7 \pm 1.9$ & 14.5\% & $6.5 \pm 7.0$ & 0.6\% \\
          $-2.8 \pm 1.3$ & 13.8\% & $-6.1 \pm 8.5$ & 0.5\% \\
          $-3.8 \pm 0.9$ & 13.6\% & $1.6 \pm 10.8$ & 0.4\% \\ 
          $1.6 \pm 2.7$ & 10.8\% & $1.8 \pm 10.5$ & 0.4\% \\
          \\[-1.8ex]
          \multicolumn{2}{c}{\obqa{}-single} & \multicolumn{2}{c}{\obqa{}-multi} \\ 
          \toprule
          $-0.9 \pm 2.7$ & 24\% & $-14.7 \pm 6.2$ & 2.4\% \\ 
          $-13.5 \pm 1.7$ & 7.8\% & $-19.4 \pm 8.5$ & 1.7\% \\ 
          $-5.8 \pm 0.7$ & 7.3\% & $-12.4 \pm 5.5$ & 1.2\% \\ 
          $8.2 \pm 5.2$ & 7.1\% & $-13.7 \pm 8.5$ & 1\% \\ 
          $3.1 \pm 1.1$ & 6.6\% & $-23.3 \pm 7.8$ & 0.8\% \\
          \\[-1.8ex]
          \multicolumn{2}{c}{\mnli{}-single} & \multicolumn{2}{c}{\mnli{}-multi} \\ 
          \toprule
          $-2.5 \pm 0.5$ & 9.6\% & $2.5 \pm 0.8$ & 1.8\% \\ 
          $-3.0 \pm 0.6$ & 4.5\% & $-1.1 \pm 0.9$ & 1.5\% \\ 
          $2.9 \pm 0.2$ & 2.8\% & $-4.6 \pm 0.8$ & 1.5\% \\ 
          $0.8 \pm 0.7$ & 2.7\% & $-1.5 \pm 0.2$ & 1.5\% \\ 
          $4.6 \pm 0.2$ & 2.4\% & $0.5 \pm 0.2$ & 1.5\%
    \end{tabular}
    \caption{Performance difference (p.d.) between single- and multi- annotator splits and random splits of identical size. Each cell shows the p.d. mean and standard deviation, as well as the development set relative size.}
    \label{tab:annotator_versus_random_splits}
\end{table}

\begin{figure}[h]
\centering
\includegraphics[scale=0.415]{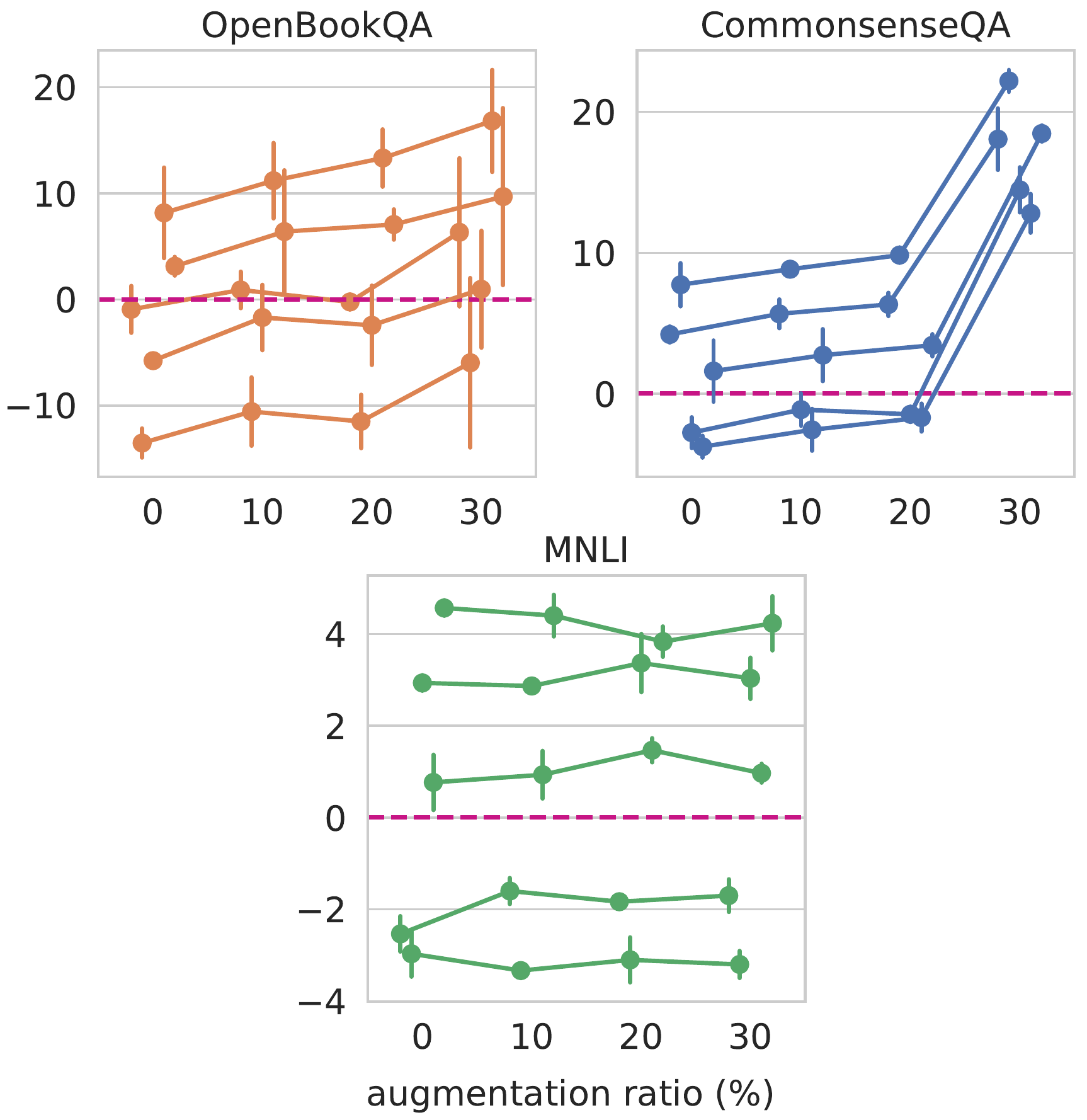}
\caption{Performance difference between single-annotator splits and random splits of identical size. The x-axis indicates the fraction of examples taken from the development set to augment the training set.}
\label{fig:augmentation}
\end{figure}

\paragraph{Separating annotator bias from annotator difficulty}

To mitigate the effect of annotator difficulty, we perform an augmentation experiment. Assume a development set $\sS_{\text{dev}}$, for which performance is low.
Our hypothesis is that if 
$\sS_{\text{dev}}$ is inherently difficult, then moving a small portion of examples from $\sS_{\text{dev}}$ to the training set $\sS_{\text{train}}$ should not change performance on $\sS_{\text{dev}}$ substantially. However, if performance on $\sS_{\text{dev}}$ is low due to annotator bias, then moving examples to $\sS_{\text{train}}$ would expose the model to the annotator and performance should go up.


For every single-annotator data split $\sS_{\text{train}}$, $\sS_{\text{dev}}$, we perform a series of augmentation experiments, where we move a random fraction of $k$ examples from $\sS_{\text{dev}}$ to $\sS_{\text{train}}$, for $k=0.1, 0.2, 0.3$. 
We keep the size of $\sS_{\text{train}
}$ constant by randomly removing examples from it before augmentation.
We repeat experiments multiple times as before, and report the p.d mean and standard deviation in Figure~\ref{fig:augmentation}.

Results for both \csqa{} and \obqa{} show a rapid increase of 10-20 accuracy point for all top-$5$ annotators, given only a small number of their generated examples. This shows that the examples generated by these annotators are not inherently difficult, and that the model can substantially improve performance by being exposed to the language that the annotators generate.
Conversely, performance changes are marginal for \mnli{}, suggesting generalization patterns are mostly due to example difficulty. 
The different results for \mnli{} compared to those observed for \obqa{} and \csqa{} may be attributed to the less-skewed annotator distribution and large number of examples in \mnli{} (see Figure~\ref{fig:example_coverage} and Table~\ref{tab:datasets}), which make the model more robust to small perturbations in the data distribution.



\section{Discussion and Conclusions}
This study set out to investigate whether prevalent crowdsourcing practices for building NLU datasets introduce an annotator bias in the data that leads to an over-estimation of model performance. We established that perfect annotator information can improve model performance, and that the language generated by annotators often reveals their identity. 
Moreover, we tested the ability of models to generalize to unseen annotators in three recent NLU datasets, and found that in two of these datasets annotator bias is evident.
These findings may be explained by the annotator distributions and the size of these datasets. Skewed annotator distributions with only a few annotators creating the vast majority of examples are more prone to biases. 

Our results suggest that annotator bias should be monitored in crowdsourcing tasks involving free text generation by annotators. 
This can be done by testing model performance on new annotators during data collection. Moreover, to tackle annotator bias, we propose that training set annotators should be separated from test-set annotators.

\section*{Acknowledgments}
This research was partially supported by
The Israel Science Foundation grant 942/16, The
Blavatnik Computer Science Research Fund and
The Yandex Initiative for Machine Learning.
We thank Sam Bowman from NYU University and Alon Talmor from Tel Aviv University for providing us the annotation information of the \mnli{} and \csqa{} datasets.
This work was completed in partial fulfillment for the Ph.D degree of the first author.

\bibliography{all}
\bibliographystyle{acl_natbib}

\end{document}